\documentclass{article} 
\usepackage{iclr2025_conference,times}

\usepackage{amsmath,amsfonts,bm}

\def\eqref#1{equation~\ref{#1}}

\def\1{\bm{1}}

\DeclareMathAlphabet{\mathsfit}{\encodingdefault}{\sfdefault}{m}{sl}
\SetMathAlphabet{\mathsfit}{bold}{\encodingdefault}{\sfdefault}{bx}{n}

\usepackage{hyperref}
\usepackage{url}
\usepackage{graphicx}
\usepackage{booktabs}
\usepackage{placeins}
\usepackage{caption}
\usepackage{subcaption}
\usepackage{tabu}
\usepackage{multirow}
\usepackage{bm}
\usepackage{amsmath}

\usepackage[utf8]{inputenc}
\usepackage[T1]{fontenc}

\definecolor{my_red}{rgb}{0.835, 0.309, 0.243}
\definecolor{my_green}{rgb}{0.333, 0.635, 0.345}
\definecolor{my_blue}{rgb}{0.298, 0.505, 0.913}
\definecolor{my_yellow}{rgb}{0.945, 0.741, 0.278}
\definecolor{my_purple}{rgb}{0.356, 0.078, 0.431}

\newlength\savewidth

\title{CrossViewDiff: A Cross-View Diffusion Model for Satellite-to-Street View Synthesis}

\author{
    Weijia Li ${^{1}}$\thanks{These authors contributed equally to this work.} \quad Jun He${^{1*}}$ \quad Junyan Ye${^{1,2*}}$ \quad Huaping Zhong${^{2,3*}}$ \\ 
  \ \textbf{Zhimeng Zheng}${^{2}}$\quad \textbf{Zilong Huang}${^{1}}$ \quad \textbf{Dahua Lin}${^{2}}$ \quad \textbf{Conghui He}${^{2,3}}$\thanks{Corresponding author(s). E-mail(s):heconghui@pjlab.org.cn}
    \\
     $^1$ Sun Yat-Sen University, China \\ $^2$ Shanghai Artificial Intelligence Laboratory, China \quad $^3$ Sensetime Research, China \\
    }

\iclrfinalcopy 
\begin{document}

\maketitle

\begin{abstract}
    
Satellite-to-street view synthesis aims at generating a realistic street-view image from its corresponding satellite-view image. 
Although stable diffusion models have exhibit remarkable performance in a variety of image generation applications, their reliance on similar-view inputs to control the generated structure or texture restricts their application to the challenging cross-view synthesis task. 
In this work, we propose CrossViewDiff, a cross-view diffusion model for satellite-to-street view synthesis. 
To address the challenges posed by the large discrepancy across views, we design the satellite scene structure estimation and cross-view texture mapping modules to construct the structural and textural controls for street-view image synthesis.
We further design a cross-view control guided denoising process that incorporates the above controls via an enhanced cross-view attention module.
To achieve a more comprehensive evaluation of the synthesis results, we additionally design a GPT-based scoring method as a supplement to standard evaluation metrics.
We also explore the effect of different data sources (e.g., text, maps, building heights, and multi-temporal satellite imagery) on this task.
Results on three public cross-view datasets show that CrossViewDiff outperforms current state-of-the-art on both standard and GPT-based evaluation metrics, generating high-quality street-view panoramas with more realistic structures and textures across rural, suburban, and urban scenes.
The code and models of this work will be released at \url{https://opendatalab.github.io/CrossViewDiff/}.

\end{abstract}

\section{Introduction}\label{sec1}

Satellite images captured by high-altitude sensors differ significantly from daily images taken by ordinary ground cameras. The overhead perspective of satellite images provides a macroscopic view that encompasses extensive regional topography, building layouts, and road networks. 
street-view images, on the other hand, are captured by mobile phones or vehicle-mounted cameras, providing a ground-level observation the scene. 
In this study, we address the task of cross-view synthesis, especially satellite-to-street view synthesis, which is an important and challenging computer vision task that has received increasing attention in recent years  \cite{shi2022geometry,Sat2Density,lu2020geometry}. 
Generating realistic street-view images from corresponding satellite images through cross-view synthesis can benefit various applications, such as cross-view geolocalization \cite{CVPR2024Unlabeled,toker2021coming}, urban building attribute recognition \cite{sg-bev}, and 3D scene reconstruction \cite{li2024sat2scene}.

\begin{figure*}[!ht]
\centering
    \includegraphics[width=\linewidth]{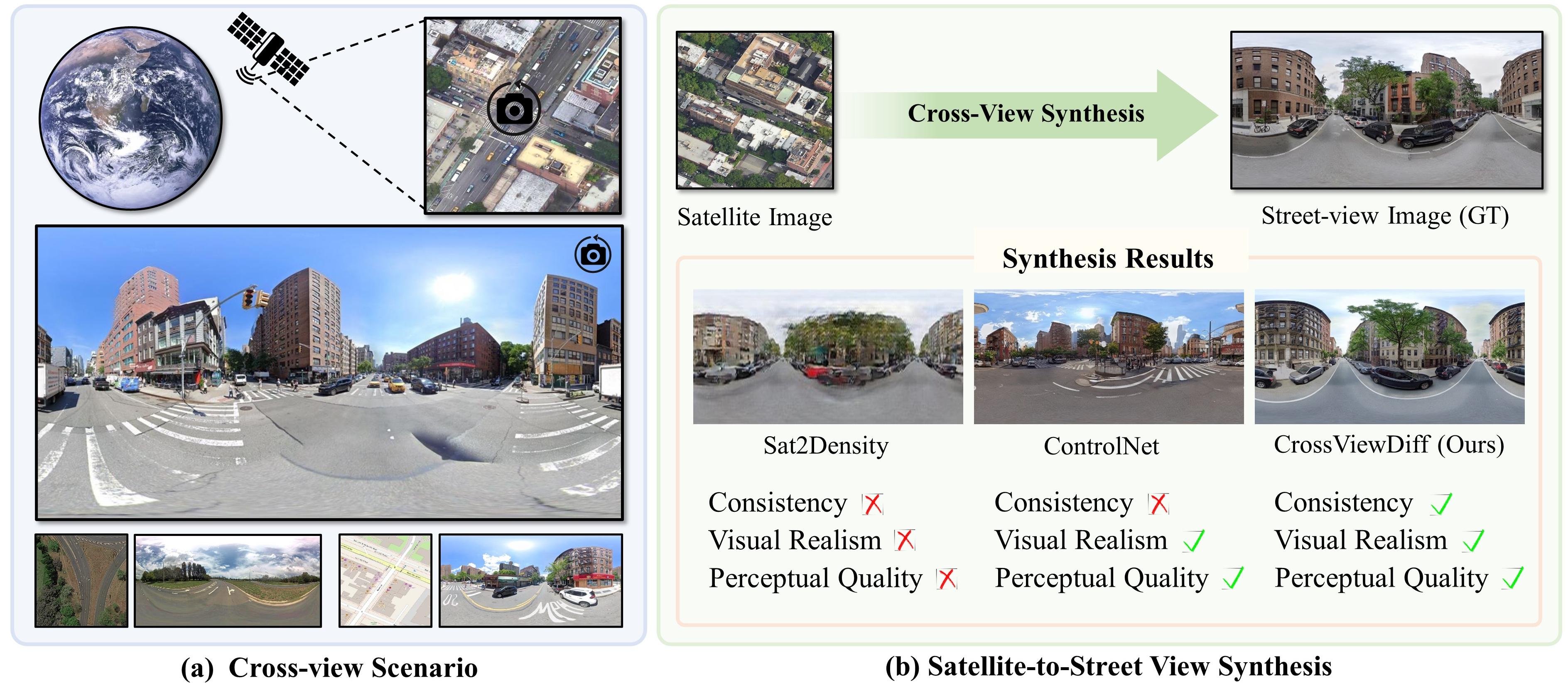}
    
    \vspace{10pt}
    \caption{
    Illustration of the satellite-to-street view synthesis task. (a) In cross-view scenarios, the satellite view and street view differ significantly, with limited overlapping information, posing a serious challenge to the satellite-to-street view synthesis task.
    (b) Compared with existing methods using GANs (e.g., Sat2Density \cite{Sat2Density}) or diffusion models (e.g., ControlNet \cite{control}), CrossViewDiff is capable of synthesizing more realistic street-view images with better perceptual quality and consistency with Ground Truth.
    }
    \label{f1}
\end{figure*}

Due to the significant differences in viewpoints and imaging methods, the overlapping information between different perspectives is very limited \cite{tang2019multi,regmi2018cross,ye2024cross,ye2024skydiffusion}, as shown in Figure \ref{f1} (a). This creates a substantial domain gap between satellite and street-view images, making the synthesis task highly challenging \cite{lu2020geometry,shi2022geometry}. 
Consequently, some studies have explored the use of additional ground truth semantic segmentation maps as auxiliary conditions for models to improve the synthesis results \cite{Zhai_2017_CVPR, regmi2018cross, tang2019multi, wu2022cross}. However, this essentially generates images from semantic maps and does not truly accomplish satellite-to-street cross-modal generation. 
Other studies have explored various satellite-to-street projection or transformation methods, utilizing geometric structure priors derived from satellite images to enhance the layout and structure of synthesized street-view panoramas  \cite{lu2020geometry,toker2021coming,shi2022geometry,Sat2Density}. 
However, there has been limited exploration of the fidelity and consistency of textures in cross-modal synthesis between satellite images and street-view panoramas.

Furthermore, existing satellite-to-street view synthesis methods are mostly based on Generative Adversarial Networks (GANs), which often result in poor image quality and unrealistic textures in the synthesized results, as shown in Figure \ref{f1} (b).

Recently, diffusion models have demonstrated superior performance in various content generation applications, garnering widespread attention \cite{ddim,ddpm,ediff,dalle2,t2i3}. Models like ControlNet enable controllable image synthesis based on various visual conditions \cite{control,huang2023composer,zhao2024uni,ruiz2023dreambooth}. For satellite-to-street view synthesis, one potential solution is to treat this task as a controllable image synthesis task, using satellite images to control the synthesis of street-view images. However, existing methods utilize similar-view images (e.g., sketches, segmentation maps) as inputs to control the structure or texture of the generated results. The different modality domains of satellite and street-view images limit the applicability of these methods in cross-view synthesis tasks. As shown in Figure \ref{f1}(b), the domain gap results in synthesized images that are often realistic yet inconsistent, with significant differences between the synthesized street-view images and the actual corresponding satellite content.

Furthermore, existing cross-view generation studies \cite{lu2020geometry,toker2021coming,regmi2018cross} commonly use image generation metrics such as SSIM \cite{SSIM2004} and PSNR to evaluate the content consistency of synthesized images, as well as FID \cite{FID} and KID \cite{KID} to assess image realism. However, these traditional metrics often fall short in aligning with human perception and lack transparency and interpretability. With the development of multimodal large models (MLLM) \cite{openai2023gpt,team2023gemini, liu2024visual, li2023blip}, an increasing number of studies have employed multimodal large models like GPT-4o \cite{openai2023gpt} for assessing the quality of synthesized images \cite{cho2023davidsonian,huang2023t2i,wu2024gpt,zhang2023gpt}, achieving interpretable and highly human-aligned scoring \cite{ku2023viescore,peng2024dreambench++}. However, prior use of multimodal scoring has predominantly been in text-to-image synthesis or editing tasks, with no studies applying it to cross-view synthesis tasks. 

In this work, we propose CrossViewDiff, a cross-view diffusion model for satellite-to-street view synthesis. Based on the geometric and imaging relationships between satellite and street views, we construct structural and texture controls from satellite images and have designed a cross-view control guided denoising process to enhance the structural and texture fidelity of synthesized panoramic images. 
Additionally, we extend the traditional satellite-to-street view synthesis task by exploring different data sources, such as text, map data, building height data and multiple-temporal satellite images. 
In our experiments, we additionally utilize GPT-4o \cite{openai2023gpt} to score synthesized street-view images as a supplement to standard metrics, aiming for a more comprehensive evaluation of the generated results. 
Experimental results demonstrate that CrossViewDiff excels on three public cross-view datasets, generating realistic and content-consistent images, showcasing outstanding synthesis quality. 

The main contributions of this work are summarized as follows:
\begin{itemize}
\item 
We design satellite scene structure estimation and cross-view texture mapping modules to overcome the significant discrepancy between satellite and street views, constructing structure and texture controls for street-view image synthesis.
 
\item We propose a novel cross-view control guided denoising process that incorporates the structure and texture controls via an enhanced cross-view attention module to achieve more realistic street-view panorama synthesis.
\item We conduct extensive experiments in street-view image synthesis across a variety of scenes (rural, suburban, and urban), explore additional data sources (e.g. text, maps, multi-temporal images, etc.), and design a GPT-based evaluation metric as a supplement to standard metrics.
\item CrossViewDiff outperforms state-of-the-art methods on three public cross-view datasets, achieving an average increase of 9.0\% in SSIM, 39.0\% in FID, and 35.5\% in the GPT-based score.

\end{itemize}

\section{Related work}
\label{sec:related_work}

\subsection{Satellite-to-street view synthesis}

Satellite-to-street view synthesis is a challenging task that has been extensively studied. To mitigate the difficulties posed by the large differences across views, many studies explored additional semantic priors to enhance the structure of street-view synthesis results \cite{Zhai_2017_CVPR,regmi2018cross,tang2019multi,wu2022cross}. Zhai et al. \cite{Zhai_2017_CVPR} is a pioneer in this domain that infers the street-view semantic map from the satellite semantic map via a learnable linear transformation. 
Tang et al. \cite{tang2019multi} utilized both the satellite image and the semantic map of street-view image as input to synthesize the target street-view image via image-to-image translation.
Although providing a strong structure prior of street-view images, the semantic map is not always available in the actual cross-view synthesis scenarios.


Another group of studies proposed satellite-to-street synthesis methods without using additional semantic information of street-view images, which explored various cross-view projection or transformation methods to provide geometry guidance specifically for panoramic image synthesis \cite{lu2020geometry,toker2021coming,shi2022geometry,Sat2Density}.
In Lu et al. \cite{lu2020geometry}, a geo-transformation method was proposed for leveraging the height map of satellite view to produce the additional building geometry condition to facilitate street-view panorama synthesis. 
Toker et al. \cite{toker2021coming} applied a polar transformation method proposed by \cite{shi2019spatial} to cross-view image synthesis and designed a multi-tasks framework in which image synthesis and retrieval are considered jointly.
Shi et al. \cite{shi2022geometry} employed a learnable geographic projection module to learn the geometry relation between the satellite and ground views to facilitate street-view panorama synthesis.
Inspired by the success of neural radiance field (NeRF) \cite{mildenhall2020nerf}, Qian et al. \cite{Sat2Density} proposed a Sat2Density that can learn a faithful 3D density field as the geometry guidance for panorama synthesis.

In summary, existing studies on satellite-to-street view synthesis are based on generative adversarial networks, with the main aim of improving the structure of synthetic image via semantic or geometric guidance, generating street-view images with low quality and unrealistic textures. 
By contrast, our study proposes a novel cross-view synthesis method based on Stable Diffusion models \cite{stable}, which designs a cross-view control guided denoising process with a novel cross-view attention module as well as structure and texture controls, generating street-view panoramas with much better perceptual quality and more realistic textures across various scenes.

\subsection{Diffusion models}
In recent computer vision studies, diffusion models \cite{ddpm} have exhibited remarkable performance in many content creation applications, such as image-to-image translation \cite{palette,bbdm}, text-to-image generation \cite{ediff,dalle2,t2i3,zhang2024open}, image enhancement \cite{enhance1,enhance2,enhance3, wang2024exploiting}, content editing \cite{edit,diffedit}, and 3D shape generation \cite{3d1,3d2,liang2024intergen, li2024instant3d}, etc.
For traditional denoising diffusion models, images are generated by progressively denoising from random Gaussian noise. For instance, Song et al. \cite{ddim} proposed denoising diffusion implicit models (DDIM) that reduce the number of denoising steps using an alternative non-Markovian formulation. In latent diffusion models (LDM) \cite{stable}, a variational autoencoder \cite{vae} is trained for compressing natural images to a latent space, where the diffusion process will be performed in later stages.

Recently, an increasing number of diffusion models have been proposed for controllable image synthesis \cite{gal2022image,control,huang2023composer,zhao2024uni,ruiz2023dreambooth}. ControlNet \cite{control} leverages both text and a variety of visual conditions (e.g., sketch, depth map, and human pose) to generate impressive controllable images, which also avoids the need to re-train the entire large model by fine-tuning pre-trained diffusion models and zero-initialized convolution layers. Composer \cite{huang2023composer} integrates global text description with various local controls to train the model from scratch on datasets with billions of samples. Uni-ControlNet \cite{zhao2024uni} enables composable control with various conditions using a single model and achieves zero-shot learning on previously unseen tasks. 
However, these methods utilize similar-view image inputs to control the structure and texture of the synthesis results, resulting in inapplicability to cross-view synthesis tasks.

In addition, several studies have proposed diffusion models for novel view synthesis tasks. For instance, MVDiffusion \cite{mv} proposes a cross-view attention module to generate consistent indoor panoramic images, and Tseng et al. \cite{consis} utilizes epipolar geometry as a constraint prior to synthesize a consistent video of novel views from a single image.
MagicDrive \cite{gao2023magicdrive} proposes a street view generation framework that leverages diverse 3D geometry controls (i.e., camera poses, road maps, and 3D bounding boxes) and textual descriptions.  
However, existing novel view synthesis methods rely on the continuity of image views or camera pose information, which cannot be satisfied in satellite-to-street cross-view settings.
Several recent studies have aimed at cross-view synthesis task via diffusion models.
Sat2Scene \cite{li2024sat2scene} proposes a novel 3D reconstruction architecture that leverages diffusion models on sparse 3D representations to directly generate 3D urban scenes from satellite imagery. 
Streetscapes \cite{deng2024streetscapes} proposes an autoregressive video diffusion framework and introduces a novel temporal interpolation approach, generating long-range consistent street-view images based on map and height data.
However, the task settings of these studies are different from the satellite-to-street view synthesis, and their methods fail to utilize satellite image information to generate realistic street-view textures.

Although diffusion models have achieved promising performance in numerous computer vision applications, few studies have been designed for the challenging satellite-to-street view synthesis task. 
In this work, we extend the application scenarios of diffusion models to satellite-to-street view synthesis. 
With both structure and texture controls from the satellite image, our cross-view guided denoising process enables the diffusion model to generate more realistic street-view panoramas.

\section{Methods}
\label{sec:methods}

\begin{figure*}[!t]
    \includegraphics[width=\linewidth]{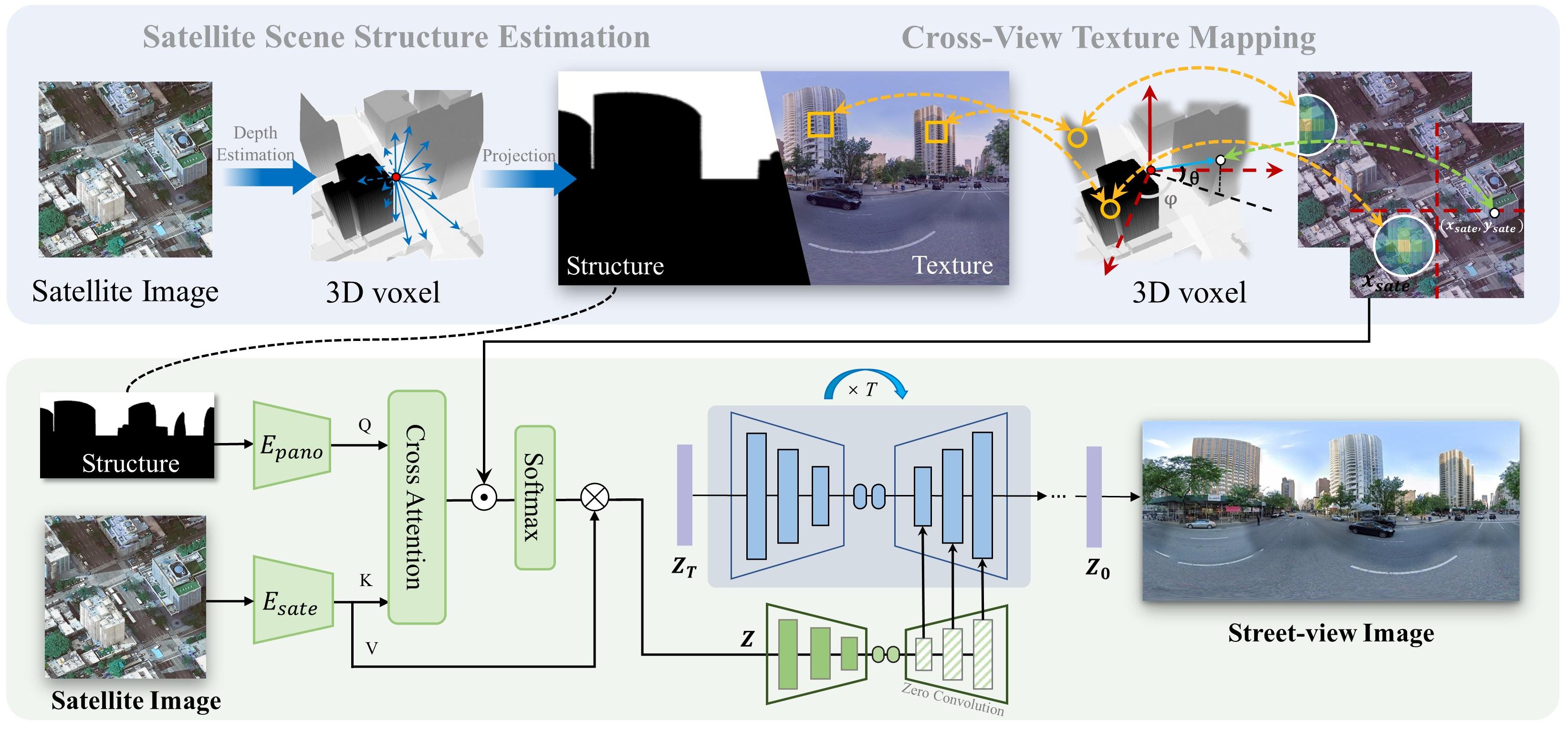}
    \caption{
     Overview of our proposed CrossViewDiff. 
     First, we create 3D voxels based on a depth estimation method as intermediaries of information across different viewpoints.
     Subsequently, based on the satellite images and 3D voxels, we establish structural and textural controls for street view synthesis via satellite scene structure estimation and cross-view texture mapping, respectively. 
     Finally, we integrate the above cross-view control information via an enhanced cross-view attention mechanism, guiding the denoising process to synthesize street-view images.     
     }
     \label{overview}
\end{figure*}

The goal of satellite-to-street view synthesis is to generate realistic and consistent street-view panoramas from corresponding satellite images. As shown in Figure \ref{overview}, this paper introduces a novel cross-view synthesis method named CrossViewDiff. In our workflow, we first construct structure and texture controls from satellite images based on the geometric and imaging relationships between satellite and street views. Subsequently, we design a cross-view control guided denoising process via an enhanced cross-view attention module, achieving the synthesis of realistic street-view images.

In the following sections, we first provide a brief introduction to the diffusion model in Section \ref{section3.1}. In Section \ref{section3.2}, we discuss the structure and texture controls for cross-view synthesis. In Section \ref{section3.3}, we describe the cross-view control guided denoising process. In Section \ref{section3.4}, we detail our strategy for effectively using the GPT model to evaluate the quality of synthesized street-view images.

\subsection{Preliminary}
\label{section3.1}
Diffusion models are generative models that can generate samples from a Gaussian distribution to match
target data distribution by a gradual denoising process \cite{ddpm}. In the
forward process, diffusion models gradually add
Gaussian noises to a ground truth image $x_0$ according to a predetermined schedule $\beta_1, \beta_2, \dots, \beta_T$:
\begin{equation}
q(x_t | x_{t-1}) = \mathcal{N}(x_t; \sqrt{1 - \beta_t}x_{t-1}, \beta_t I) 
\end{equation}
where $x_t$ is a noised sample with noise level $t$. The reverse process involves a series of denoising steps, where noise is progressively removed by employing a neural network $\epsilon_\phi$ with parameters $\phi$. This neural network predicts the noise $\epsilon$ present in a noisy image $x_t$ at step $t$. The simplified version of the loss function for training the diffusion model is formulated as follows:
\begin{equation}
L_{simple}(\phi,x) = E_{t,\epsilon}\left[\left\| \epsilon_\phi(x_t, t) - \epsilon \right\|^2\right]
\end{equation}
where $t$ is uniformly sampled from the set $\{1, \ldots, T\}$, and $x_{t-1}$ can be reconstructed from $x_t$ by removing the predicted noise:

\begin{equation}
x_{t-1} = \frac{1}{\sqrt{\alpha_t}}\left(x_t - \frac{1 - \alpha_t}{\sqrt{1-\bar{\alpha}_t}}\epsilon_\phi(x_t, t)\right) + \sqrt{\beta_t}\epsilon
\end{equation}
where $\alpha_t = 1 - \beta_t$, $\bar{\alpha}_t$ is the cumulative sum of $\alpha_t$  and $\epsilon \sim \mathcal{N}(0, I)$.

\subsection{Structure and Texture Controls for Cross-View Synthesis}
\label{section3.2}

To precisely control the generation of panoramas in cross-view scenarios, it is essential to establish structural and textural information from a street-view perspective based on satellite imagery. Specifically, we start by constructing three-dimensional voxels as intermediaries from the depth estimation results of satellite images. The structural control information is derived from projecting these 3D voxels onto the street-view panorama to obtain scene structure estimates. On the other hand, texture control is achieved through a weight matrix derived from the cross-view mapping relationship based on 3D voxels, representing the response regions on the street view image to different features of the satellite image.

\subsubsection{Satellite Scene Estimation for Structure Control}

Considering the substantial differences in viewing angles between satellite and street-view modalities, directly extracting contour information from satellite images is challenging. Therefore, we first utilize depth estimation methods to obtain depth results from the satellite perspective \cite{fu2018dorn,Chen2019SARPN,depthanything, ke2024repurposing}. Following this, we convert these depth results into a 3D voxel grid, which serves as an intermediary for scene structure reconstruction. Finally, leveraging the equiangular projection characteristics of street-view panoramas, we establish a mapping from the 3D voxels to the central street view \cite{lu2020geometry}, resulting in a binary map that represents structural information, as shown in Figure \ref{overview}. This structural information, which includes the positional distribution of significant features (such as buildings, trees, roads, etc.), will further be used as structural control in our diffusion model.

\subsubsection{Cross-View Mapping for Texture Control}

Previous methods typically utilize the global texture information of satellite images for panorama synthesis. In contrast, we propose Cross-View Texture Mapping (CVTM), which achieves localized texture control by computing the mapping relationship between each coordinate of the panorama and the satellite image.
Based on the 3D voxel grid, we calculate the elevation \( \theta \) and azimuth \( \phi \) angles from the panoramic image coordinates. For a pixel at \( (x_{\text{pano}}, y_{\text{pano}}) \) in the panoramic image, the angles are determined as follows:

\begin{align}
    \theta &= \frac{\pi}{2} - \frac{y_{\text{pano}} \cdot \pi}{\hat{H}_{\text{pano}}} \\
    \phi &= \frac{x_{\text{pano}} \cdot 2\pi}{\hat{W}_{\text{pano}}} - \pi
\end{align}

Here \( \hat{H}_{\text{pano}} \) and \( \hat{W}_{\text{pano}} \) denote the height and width of a panoramic image. The calculated angles, \(\theta\) and \(\phi\), fall within the range \([-\frac{\pi}{2}, \frac{\pi}{2}]\) and \([-\pi, \pi]\), respectively.
According to the two calculated angles, we can determine a ray starting from the center coordinate \( (x_{\text{cen}} \),  \( y_{\text{cen}}) \) of the 3D voxel map.
The length of the ray \(R\) is the distance from its first intersection with the 3D voxel grid to the center coordinate. 
Based on the above information, the final mapping coordinates in the satellite image are calculated as follows:

\begin{align}
    x_{\text{sate}} &= x_{\text{cen}} + R \cdot \cos(\theta) \cdot \cos(\phi) \\
    y_{\text{sate}} &= y_{\text{cen}} - R \cdot \cos(\theta) \cdot \sin(\phi)
\end{align}
Consequently, we establish the pixel-wise mapping relation between each panoramic coordinate \( (x_{\text{pano}}, y_{\text{pano}}) \) and its corresponding satellite-view coordinate \( (x_{\text{sate}} \), \( y_{\text{sate}}) \).

In addition, considering the intrinsic errors in cross-view alignment and other factors in complex real-world environment, it is not enough to rely on one-to-one mapping to supplement texture information (the green arrow in Fig \ref{overview}). The pixels around the mapped points in the satellite images are also valuable texture references that we need to exploit.
Consequently, we further design an enhanced satellite texture mapping strategy that leverages the surroundings of the mapped points in the satellite image to enhance the texture details in the street-view image (the orange arrows in Fig \ref{overview}). This technique utilizes an adaptive re-weighting mechanism based on the distance between the mapped point and other pixels in the satellite image. The values in the weight matrix are calculated as follows:

\begin{equation}
    M_j = 1 - \text{sigmoid}\left(\beta \left(\lVert \mathbf{p}^{*} - \mathbf{p}_{j} \rVert_2\right)\right)
\end{equation}

In this formula, \( \mathbf{p}^{*} \) indicates the coordinate \( (x_{\text{sate}} \), \( y_{\text{sate}}) \) in the satellite image that is mapped from the street-view image according to formula (4)-(7). 
The \( \mathbf{p}_{j} \) represents each pixel position in the satellite image, where \( j \) is an index ranging in \( j \) \(\in  [1, N] \), and \( N \) is the number of pixels in the satellite image. The term \( \lVert \cdot \rVert_2 \) is the Euclidean distance. The parameter \( \beta \) controls the rate of change in the sigmoid function. 
The weight value \( M_j \) indicates the importance of \(\mathbf{p}_{j}\) to the mapped point \( \mathbf{p}^{*} \), which will be higher if \(\mathbf{p}_{j}\) is close to \( \mathbf{p}^{*} \), thus enhancing the overall realism and coherence of the street-view images. Consequently, we have obtained the weight matrix $M$, which reflects the texture mapping relationship between satellite and street-view images.

\subsection{Cross-View Control Guided Denoising Process}
\label{section3.3}

Based on satellite scene estimation, we obtain binary maps to serve as structural controls for the street-view images. Utilizing cross-view mapping, we derive weight matrices to act as texture controls for the street-view images. Based on the characteristics of the structural and textural control information, we design an enhanced cross-view attention module to integrate both types of information, guiding the subsequent denoising process. 

In our enhanced cross-view attention module, let \( Q \in \mathbb{R}^{h_p \times w_p} \) denote the Query feature from the panoramic binary map $S_{pano}$, \( K \in \mathbb{R}^{h_s \times w_s} \) denote the Key feature from the input satellite image $I_{sate}$, and \(V \in \mathbb{R}^{h_s \times w_s} \) denote the Value feature, which contain detailed texture information from the satellite image. Here, \( h_p \times w_p \) and \( h_s \times w_s \) represent the resolution of the panorama and satellite feature map, respectively. 
Moreover, \(E_{\text{sate}}\) and \(E_{\text{pano}}\) denote the satellite and panoramic encoders. \(W_q\), \(W_k\) and \(W_v\) are projection matrices.
The definitions of \( Q, K, V \) are as follows:

\begin{equation}
Q = W_q(E_{\text{pano}}(S_{\text{pano}})), \quad K = W_k(E_{\text{sate}}(I_{\text{sate}})), \quad V = W_v(E_{\text{sate}}(I_{\text{sate}})),
\end{equation}

The process begins with the computation of an affinity matrix \( A \in \mathbb{R}^{h_pw_p \times h_sw_s} \), reflecting the interaction between \( Q \) and \( K \).
Following this, the weight matrix derived from the previous module is down-sampled to \( M  \in \mathbb{R}^{h_pw_p \times h_sw_s}\) and applied to each pixel within the satellite image to emphasize relevant features. 
This selective enhancement is crucial for the subsequent fusion of the detailed texture information from the satellite image into the panoramic feature \( F_{\text{pano}} \in \mathbb{R}^{h_p \times w_p} \). The enhanced cross-view attention mechanism is formulated as follows:

\begin{equation}
\text{z} = \text{softmax}(A \odot M) \cdot  V
\end{equation}
In these expressions, \( \odot \) denotes element-wise multiplication, where the weight matrix \( M \) is applied to the affinity matrix \((A)\) to obtain the reweighted affinity matrix \((A')\), emphasizing the connection between the most relevant pixels.

The output \( \text{z} \), generated at each step, is ingeniously reincorporated into the network as a pivotal conditional element. By employing \( \text{z} \) as a dynamic conditional catalyst within our cross-view diffusion architecture, we ensure that each step of the denoising process is informed by the evolving latent representation, thereby enabling a controlled and gradual transition from \(z_t\) to \(z_0\). This process is meticulously orchestrated by a cross-view control guided denoising process, which integrates structural and textural knowledge extracted from \(I_{\text{sate}}\) into the refinement of the final latent feature \(z\), subsequently decoded through Stable Diffusion's latent space decoder \(\mathcal{D}\) to achieve the generated street-view panorama \(I_{\text{pano}}\).

\subsection{GPT-based evaluation method for Cross-View Synthesis}
\label{section3.4}
\begin{figure*}[!t]
    \includegraphics[width=\linewidth]{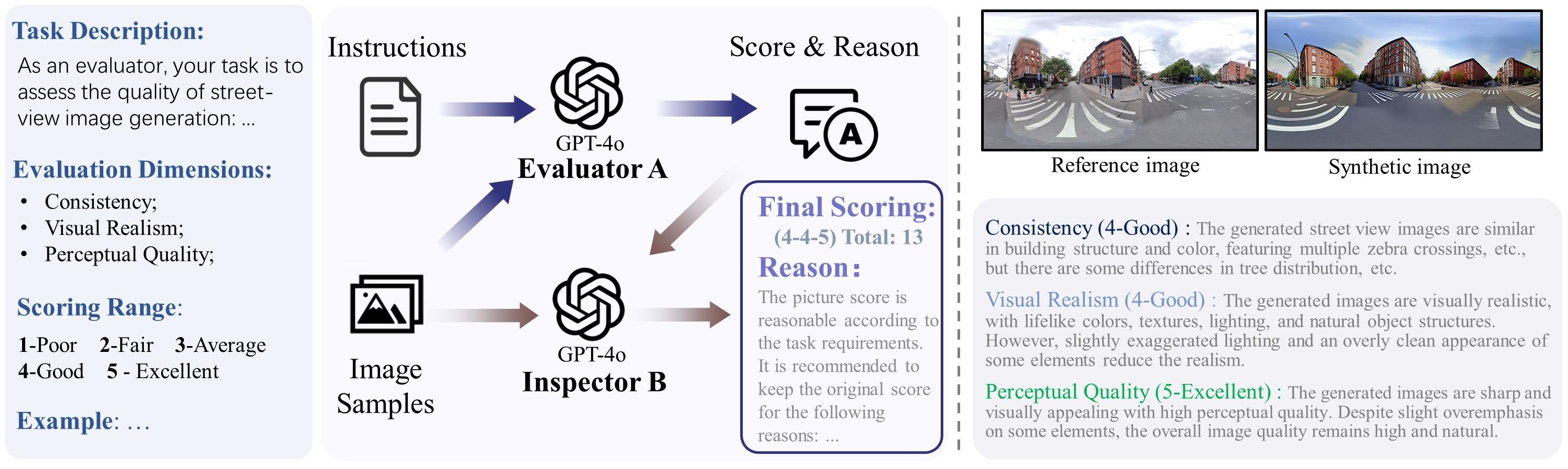}
    \caption{
    The overall process for automated evaluation using GPT-4o. 
    Instructions are meta-prompts that include a task description, scoring criteria, scoring range, and scoring examples. 
    Then we use a GPT-4o as Evaluator A to provide initial scores and reasons based on the input prompts and image samples. 
    Finally, the scores are combined with the image samples for a secondary evaluation by another GPT-4o as Inspector B, who assesses the score's appropriateness and determines the final score.
    }
    \label{GPT-score}
\end{figure*}

Cross-modal satellite-to-ground synthesis requires measuring both the consistency and realism of generated images. Traditional metrics like SSIM \cite{SSIM2004} and FID \cite{FID} generally focus on single dimensions of similarity or realism, providing incomplete evaluations. Inspired by the use of large multimodal models for synthetic image scoring \cite{cho2023davidsonian, huang2023t2i, wu2024gpt, zhang2023gpt}, we design a new evaluation process based on GPT-4o, as shown in Figure \ref{GPT-score}. This approach enables comprehensive and interpretable assessments of synthesized street-view images, aligning more closely with human judgment standards.

Firstly, we design three key evaluation dimensions for cross-view synthesized images: Consistency, Visual and Structural Realism, and Perceptual Quality. We adopted a rating scheme, establishing a 5-level rating system with scores ranging from 1 (poor) to 5 (excellent).

\textbf{Consistency:} This dimension evaluates the alignment of the content in synthesized images with real street-view images, including the structure and texture of buildings, the layout of roads, and the similarity of other significant landmarks, measuring the content consistency of the synthesized street-view images.

\textbf{Visual Realism:} This evaluates the visual effect and structural reasonableness of the generated images, including the realism of color, shape, and texture, as well as the the structural integrity, assessing whether they look like real street-view images.

\textbf{Perceptual Quality:} This evaluates the overall perceptual quality of the generated images, including aspects such as image clarity, noise level, and visual comfort, measuring the quality of the generated images.

To achieve more effective GPT scoring, we employed Chain-of-Thought (CoT) and In-Context Learning (ICL) strategies \cite{alayrac2022flamingo, zhang2023multimodal, brown2020language, peng2024dreambench++} to enhance its stability and effectiveness. Firstly, we provided GPT-4o with a small number of effective human-scored examples from multiple users, enabling the model to effectively learn human scoring patterns. Secondly, by enabling the large model to explain the reasoning behind its scores, we have introduced an element of internal reflection to the evaluation process. Additionally, we used GPT-4o to act as Evaluator A and Inspector B. After receiving the initial scores and reasons from Evaluator A, Inspector B will assess the reasonableness of these scores and make the final scoring decision. If the scores are deemed reasonable, they will be retained; otherwise, Inspector B will provide new scoring results and justifications.


To validate the effectiveness of GPT-based scoring, we invited ten human users to perform the same scoring task and measured the consistency between their scores and those generated by GPT. We provided thorough training to the users to ensure they fully understood the satellite-to-street view generation task. The users' scoring tasks and schemes were aligned with the GPT scoring. We ensured that each generated image was scored by at least two human users. Due to the large volume of cross-view datasets and the cost of both user and GPT scoring, we randomly sampled 1000 images from the evaluation sets of each dataset for assessment. In addition to our method, we selected the best comparative results from GAN and diffusion methods for evaluation. A total of 9000 images were used for user scoring, and we measured the agreement between these scores and the GPT scores.

\section{Experiments}
\label{sec:experiments}

In this section, we first introduce the three datasets used in this study and the experimental setting. Next, we conduct both qualitative and quantitative comparisons of CrossViewDiff with state-of-the-art cross-view synthesis methods. Following this, we perform ablation studies to evaluate the effectiveness of each module. Additionally, we explore street-view synthesis tasks using additional data sources. Finally, we discuss the limitations of our method.

\subsection{Dataset}

In our experiments, we used three popular cross-view datasets to evaluate the synthesis results, i.e., CVUSA \cite{Zhai_2017_CVPR}, CVACT \cite{CVACT} and OmniCity \cite{Li_2023_CVPR}. These three datasets encompass rural, suburban, and urban scenes, providing a robust benchmark for comprehensively evaluating the performance of satellite-to-street view synthesis. Furthermore, in addition to the original satellite imagery and building height data provided by OmniCity, we supplemented multimodal data including text, maps, and multi-temporal satellite imagery, providing data support for street-view synthesis tasks using additional multimodal data sources.

\textbf{CVUSA \cite{Zhai_2017_CVPR} }is a standard large-scale cross-view benchmark, primarily featuring rural scenes such as roads, grasslands, and forests. This dataset comprises centrally aligned satellite and street-view images collected from various locations across the United States, which is randomly split into training and test sets in an 8:2 ratio.

\textbf{CVACT \cite{CVACT}} is a widely used cross-view dataset that includes satellite and street-view images from Canberra, Australia. This dataset mainly consists of suburban scenes with relatively low buildings and open views. Unlike CVUSA dataset, the training and test sets of CVACT dataset are divided by region. 

%

\textbf{OmniCity \cite{Li_2023_CVPR} }is an urban cross-view dataset that includes street-view and satellite images from New York, USA. The primary scenes in OmniCity consist of dense urban buildings, and street-view images that are heavily obstructed by trees or vehicles will be filtered out. OmniCity is divided into training and test data by region.

Additionally, the orientation towards the north in both street view and satellite imagery is a critical attribute for cross-view datasets. In all three datasets, the north direction in satellite images is at the top of the image. In CVUSA \cite{Zhai_2017_CVPR} and CVACT \cite{CVACT}, the north direction in street-view images is in the center column, while in OmniCity \cite{Li_2023_CVPR}, it is in the first column. 

\subsection{Experimental Setting}

We implement CrossViewDiff based on the ControlNet \cite{control} framework, incorporating the pre-trained Stable Diffusion \cite{stable} v1.5 model. The diffusion decoder is configured in an unlocked state and the classifier-free guidance \cite{cfg} scale is established at 9.0. For the final inference sampling, we adopt $T=50$ as the sampling step, consistent with the DDIM \cite{ddim} strategy. The entire training process is performed on eight NVIDIA A100 GPUs, with a batch size of 128, spanning a total of 100 epochs. 
Our depth estimation method employs Marigold \cite{ke2024repurposing} and is fine-tuned on the OmniCity dataset, which provides elevation information. 
We conduct our experiments at a resolution of $1024 \times 256$ on the CVUSA \cite{Zhai_2017_CVPR} and $1024 \times 512$ on OmniCity \cite{Li_2023_CVPR} and CVACT \cite{CVACT}.

We compared our method with several state-of-the-art cross-view synthesis methods on the three datasets, including GAN-based methods such as Sate2Ground \cite{lu2020geometry}, CDTE \cite{toker2021coming}, S2SP \cite{shi2022geometry}, and Sat2Density \cite{Sat2Density}, as well as diffusion models for image transformation control like ControlNet \cite{control} and Instruct pix2pix (Instr-p2p) \cite{insd}. 
For Sat2Density \cite{Sat2Density}, we follow their original setup, i.e., the lighting hints are determined based on the average values of the sky histograms obtained from random selections. For diffusion-based methods (ControlNet and instr-p2p), we use a pre-trained model consistent with that of CrossViewDiff and maintain the same sample steps. 
Note that all comparison methods are conducted according to their optimal experimental settings.

Following previous studies \cite{lu2020geometry,toker2021coming,regmi2018cross}, we used common metrics such as SSIM \cite{SSIM2004}, SD, and PSNR to evaluate the content consistency of synthesized images, and FID \cite{FID} and KID \cite{KID} to assess image realism. Furthermore, in Section \ref{section4.3.2}, we use GPT-4o to evaluate the synthesized street view images across three dimensions: consistency, visual realism, and perceptual quality.

\subsection{Comparison with State-of-the-art methods}

\subsubsection{Quantitative and Qualitative Evaluation}

\begin{table*}[!h]
\centering
\caption{
Quantitative comparison of different methods on CVUSA \cite{Li_2023_CVPR} and CVACT \cite{CVACT} datasets in terms of various evaluation metrics. 
}

\scalebox{0.64}{%
\begin{tabular}{l|ccccc|ccccc}
\toprule
\multirow{2}{*}{Method} & \multicolumn{5}{c|}{CVUSA}  & \multicolumn{5}{c}{CVACT}\\
\cmidrule(lr){2-6} \cmidrule(lr){7-11}
 & SSIM ($\uparrow$)  & SD ($\uparrow$)  & PSNR ($\uparrow$)& FID ($\downarrow$) & KID ($\downarrow$)  & SSIM ($\uparrow$) & SD ($\uparrow$) & PSNR ($\uparrow$)  & FID ($\downarrow$)& KID ($\downarrow$) \\
\midrule
Sate2Ground \cite{lu2020geometry}& 0.294 & 15.48 & 12.634 & 52.42   & 0.036 & 0.392 & 15.09  & 13.038 & 55.61 & 0.079 \\
CDTE \cite{toker2021coming}  & 0.283 & 15.24 & 13.815  & 28.35 
 & 0.028  & 0.370 & 15.52 & 13.707 & 57.00 & 0.064 \\
S2SP \cite{shi2022geometry}  & 0.319 & 15.73  & 13.689   & 27.31 & 0.021 & 0.368 & 15.86 & 13.974 & 65.38 & 0.064 \\
Sat2Density \cite{Sat2Density}  & 0.339  & 15.73 & \textbf{14.229} & 41.43  & 0.036 & 0.387 & 16.09 & \textbf{14.271} & 47.09 & 0.038 \\
ControlNet \cite{control} & 0.277 & 15.22  & 11.182 & 44.63 & 0.044 & 0.340 & 15.36 & 12.150 & 47.15 & \textbf{0.019} \\
Instruct pix2pix \cite{insd}  & 0.255 & 15.76  & 10.664  & 68.75 & 0.077 & 0.392 & 15.64 & 13.123 & 57.74 & 0.049 \\
\midrule
Ours   & \textbf{0.371}  & \textbf{16.31} & 12.000 & \textbf{23.67} & \textbf{0.018} & \textbf{0.412} & \textbf{16.29} &  12.411 & \textbf{41.94} & 0.041 \\
\bottomrule
\end{tabular}
}
\label{tab:method_comparison0}
\end{table*}

\begin{figure*}[!h]  
    \centering
    \includegraphics[width=\linewidth]{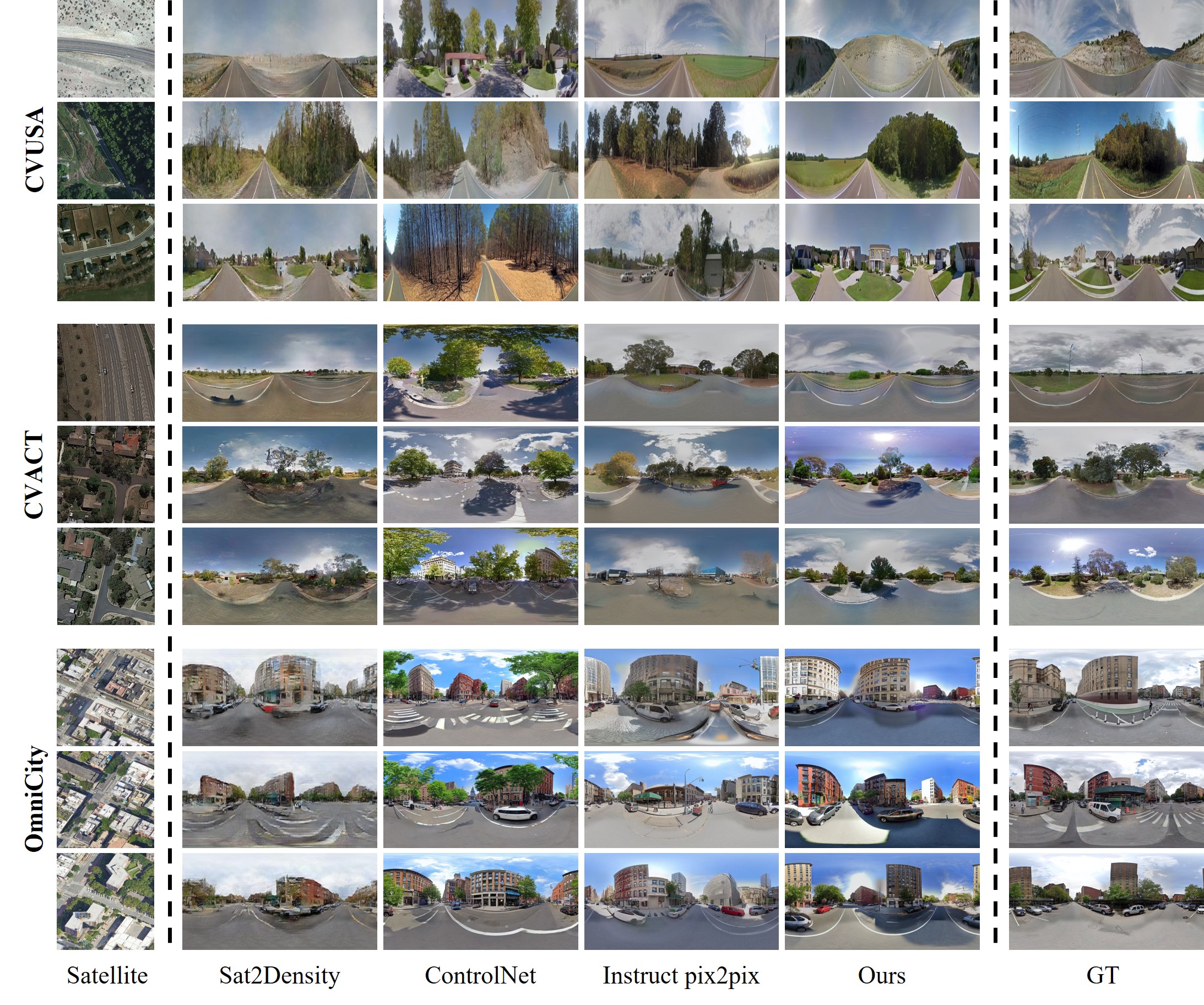}
    \caption{Qualitative comparison of synthesis results on CVUSA \cite{Zhai_2017_CVPR}, CVACT \cite{CVACT} and OmniCity \cite{Li_2023_CVPR}, respectively. The comparison includes the synthesis results of Sat2Density \cite{Sat2Density}, ControlNet \cite{control}, Instr-p2p \cite{insd}, and our method. The results indicate that our method generates street views that are more realistic, consistent, and of higher quality compared to other methods.}
    \label{fig:comparison}
\end{figure*}

\begin{table*}[!h]
\centering

\caption{
Quantitative comparison of different methods on  OmniCity \cite{Li_2023_CVPR} dataset in terms of various evaluation metrics. 
}
\scalebox{0.75}{%
\begin{tabular}{l|ccccc}
\toprule
\multirow{2}{*}{Method} & \multicolumn{5}{c}{OmniCity} \\
\cmidrule(lr){2-6} 
 & SSIM ($\uparrow$)  & SD ($\uparrow$)  & PSNR ($\uparrow$)& FID ($\downarrow$) & KID ($\downarrow$) \\
\midrule
Sate2Ground \cite{lu2020geometry}& 0.290 & 14.38 & 12.430 & 75.22   & 0.053 \\
CDTE \cite{toker2021coming}  & 0.294 & 14.47 & 11.594  & 122.29 & 0.141 \\
S2SP \cite{shi2022geometry}  & 0.294 & 14.61  & 12.748   & 84.00 & 0.088  \\
Sat2Density \cite{Sat2Density}  & 0.316  & 14.73 & \textbf{13.661} & 87.90  & 0.072 \\
ControlNet \cite{control} & 0.297 & 14.64  & 10.703 & 59.99 & 0.056 \\
Instruct pix2pix \cite{insd}  & 0.291 & 14.03  & 10.363  & 64.89 & 0.087 \\
\midrule
Ours   & \textbf{0.353}  & \textbf{15.17} & 11.127 & \textbf{42.01} & \textbf{0.033} \\
\bottomrule
\end{tabular}
}
\label{tab:method_comparison1}
\end{table*}

We provide the quantitative results on the rural CVUSA and suburban CVACT datasets in Table \ref{tab:method_comparison0}.
Compared to the state-of-the-art method for cross-view synthesis (Sat2Density), our method achieved significant improvements in SSIM \cite{SSIM2004} and FID \cite{FID} scores by 9.44\% and 42.87\% on CVUSA, respectively. Similarly, enhancements of 6.46\% and 10.94\% in SSIM and FID were observed on CVACT. Visual results from Figure \ref{fig:comparison} suggest that GAN-based cross-view methods tend to produce excessive artifacts and blurriness. 
While diffusion-based approaches like ControlNet \cite{control} and Instr-p2p \cite{insd} can generate highly realistic street views, they often lack content relevancy with the Ground Truth. In contrast, our method benefits from structure and texture controls, effectively capturing satellite-view information to generate realistic images that are more consistent with the Ground Truth street-view images, including buildings, trees, green spaces, and roads.

In the urban OmniCity dataset, our CrossViewDiff also demonstrates excellent performance compared to the most advanced methods, as shown in Table \ref{tab:method_comparison1}. Compared with the state-of-the-art (Sat2Density \cite{Sat2Density}), our approach achieves significant improvements in SSIM \cite{SSIM2004} and FID \cite{FID} by 11.71\% and 52.22\%, respectively. The visual results from the last three rows of Figure \ref{fig:comparison} demonstrate that our method effectively maintains good performance in synthesized street view images of urban scenes, such as more realistic and consistent building contours and colors.
Extensive experimental results demonstrate that our CrossViewDiff outperforms existing methods and achieves excellent results for street-view image synthesis across various scenes, including rural, suburban and urban environments.

\subsubsection{GPT-based Evaluation}
\label{section4.3.2}

\begin{figure*}[ht]
    \includegraphics[width=\linewidth]{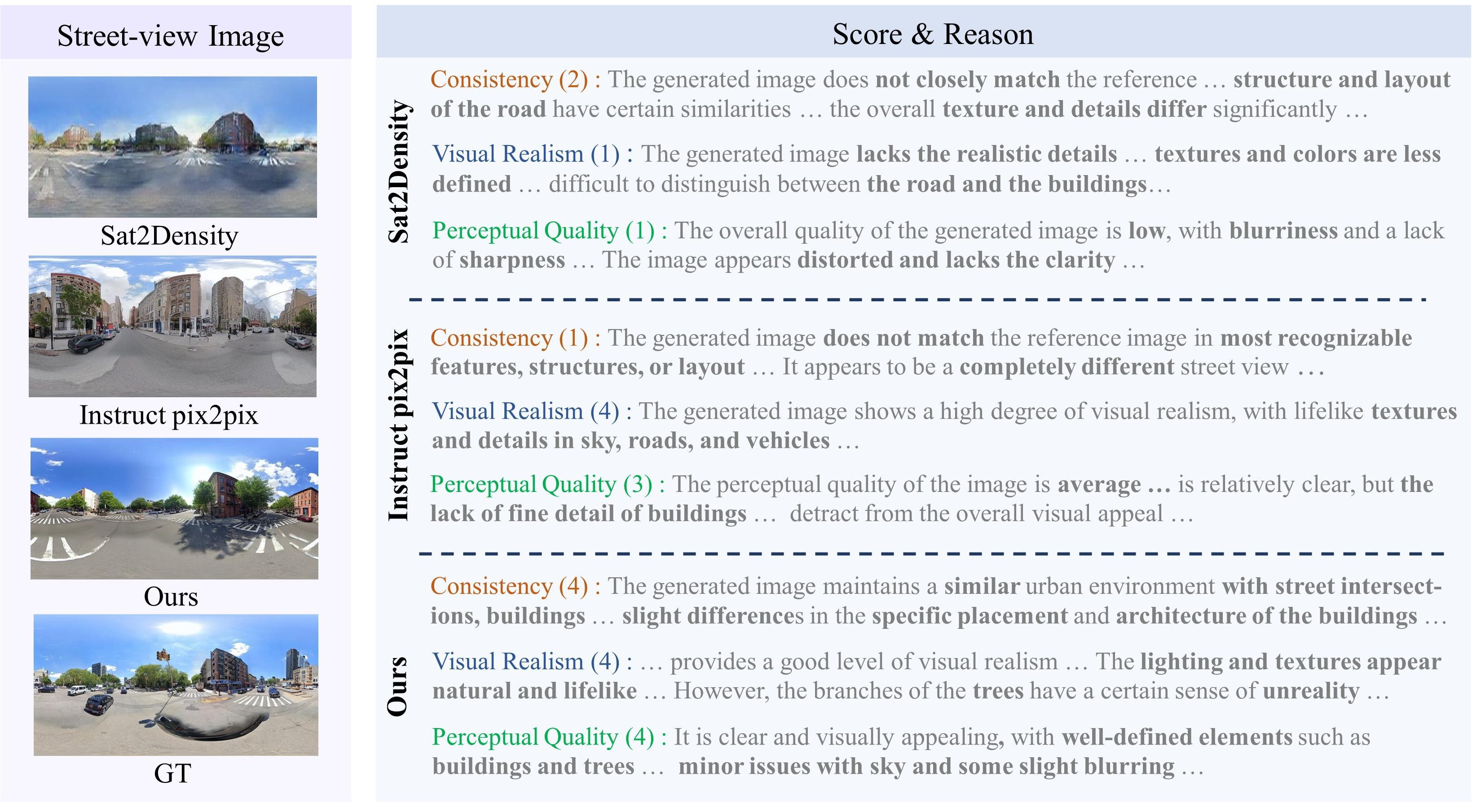}
    \caption{An example of GPT-based evaluation. Given a synthesized street-view image and the corresponding Ground Truth, GPT-based evaluation can provide scores  across multiple dimensions and the corresponding reasons for the scores.
    }
    \label{GPT_Score_Case}
\end{figure*}

\begin{figure}[ht]
\begin{minipage}[t]{0.48\linewidth}
\vspace{0pt}
\centering
\captionof{table}{Average similarity between human user ratings and GPT ratings. }
\begin{tabular}{l|c}
\toprule
Evaluation Metrics & Average Similarity \\
\midrule
Consistency & 0.810  \\
Visual Realism  & 0.816  \\
Perceptual Quality  & 0.743 \\
Total Score  & 0.801  \\
\bottomrule
\end{tabular}
\label{tab:GPT-human}
\end{minipage}
\hfill
\begin{minipage}[t]{0.48\linewidth}
\vspace{0pt}
\centering
\caption{GPT-based evaluation results.}
\includegraphics[width=0.8\linewidth]{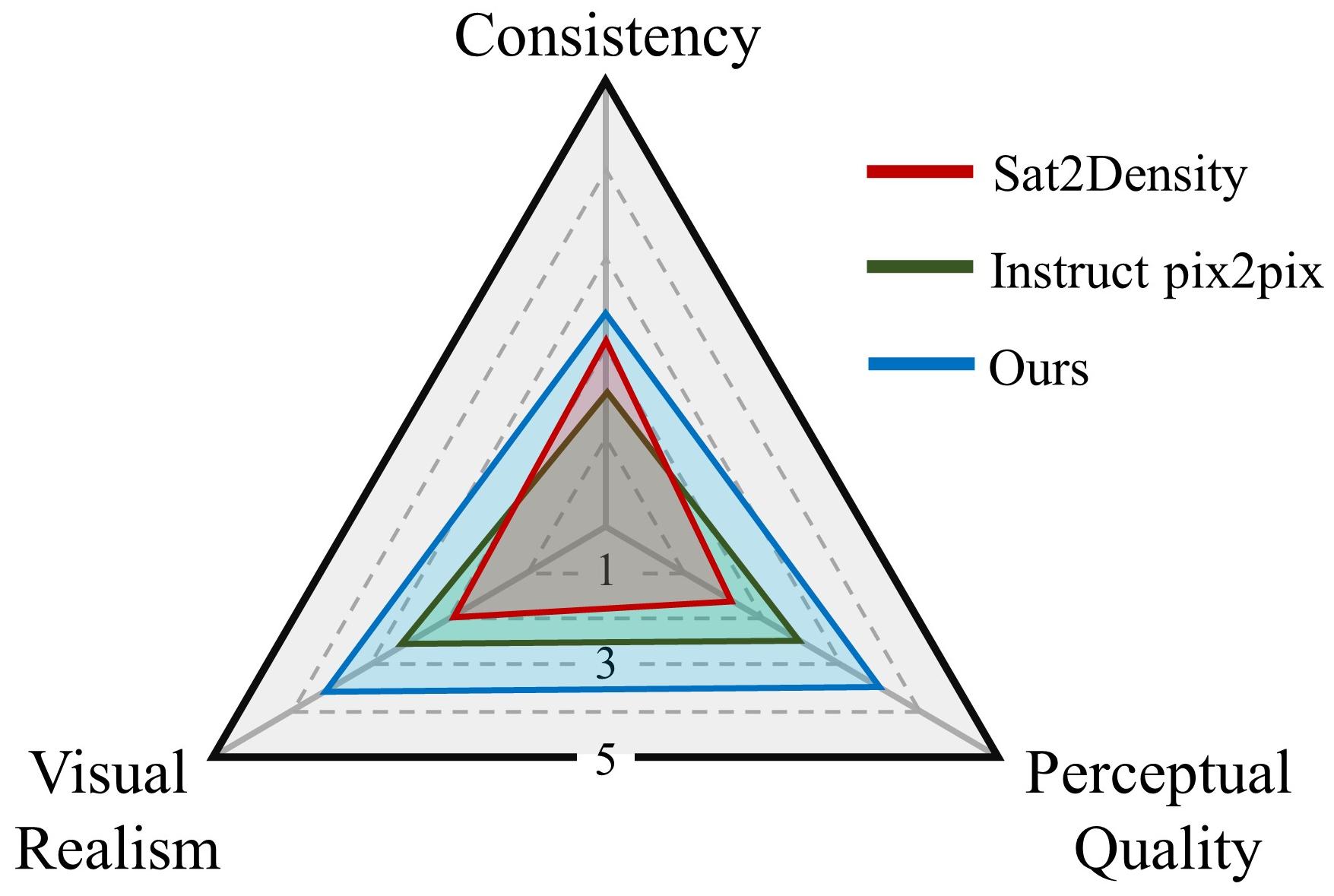}
\label{GPT-lidar}
\end{minipage}
\end{figure}

Beyond conventional similarity and realism metrics, we also leverage the powerful visual-linguistic capabilities of existing MLLM large models to design a GPT-based scoring method for evaluating synthetic images. As shown in Figure \ref{GPT_Score_Case}, GPT can provide scores across multiple dimensions along with the corresponding reasons for the scores. The description of the scoring reasons by GPT enhances the interpretability of the metric scores. As described in section \ref{section3.4}, a subset of the dataset (9K pairs of images) was evaluated by both human users and GPT. By calculating the similarity between each user rating and the GPT score, the results, as shown in Table \ref{tab:GPT-human}, demonstrate that GPT-based scoring performs well in aligning with human ratings across multiple metrics, with an average similarity exceeding 80\%. 
This highlights the fact that GPT-based scoring is very close to human preferences and can effectively evaluate synthetic street-view images.

Moreover, as illustrated in Table \ref{tab:Cross-score} and Figure \ref{GPT-lidar}, our method significantly outperforms other GAN-based and diffusion-based generation methods in the three evaluation dimensions of Consistency, Visual Realism, and Perceptual Quality. This also indicates that the street-view images synthesized by our method are more aligned with the requirement of human users, which aids in subsequent applications such as immersive scenes and virtual reality tasks.

\begin{table*}[!h]
\centering
\caption{
Evaluation results of street view synthesis based on GPT-4o. The scores range from 1 (poor) to 5 (excellent), presenting the average score across three datasets.
Our method significantly outperforms other methods in terms of the three evaluation dimensions and the total score.
}

\scalebox{0.95}{%
\begin{tabular}{l|cccc}
\toprule
 Method & Consistency  & Visual Realism & Perceptual Quality  & Total Score \\
\midrule
Sat2Density \cite{Sat2Density} & 2.07 & 2.05 & 1.74  & 7.91 \\
Instruct pix2pix \cite{insd}  & 1.67 & 2.75  & 2.61  & 9.79 \\
Ours  & \textbf{2.32} & \textbf{3.66}   & \textbf{3.66} & \textbf{13.27}  \\
\bottomrule
\end{tabular}
}
\label{tab:Cross-score}
\end{table*}

\subsubsection{Panorama Continuity Evaluation}

\begin{figure*}[ht]
    \includegraphics[width=\linewidth]{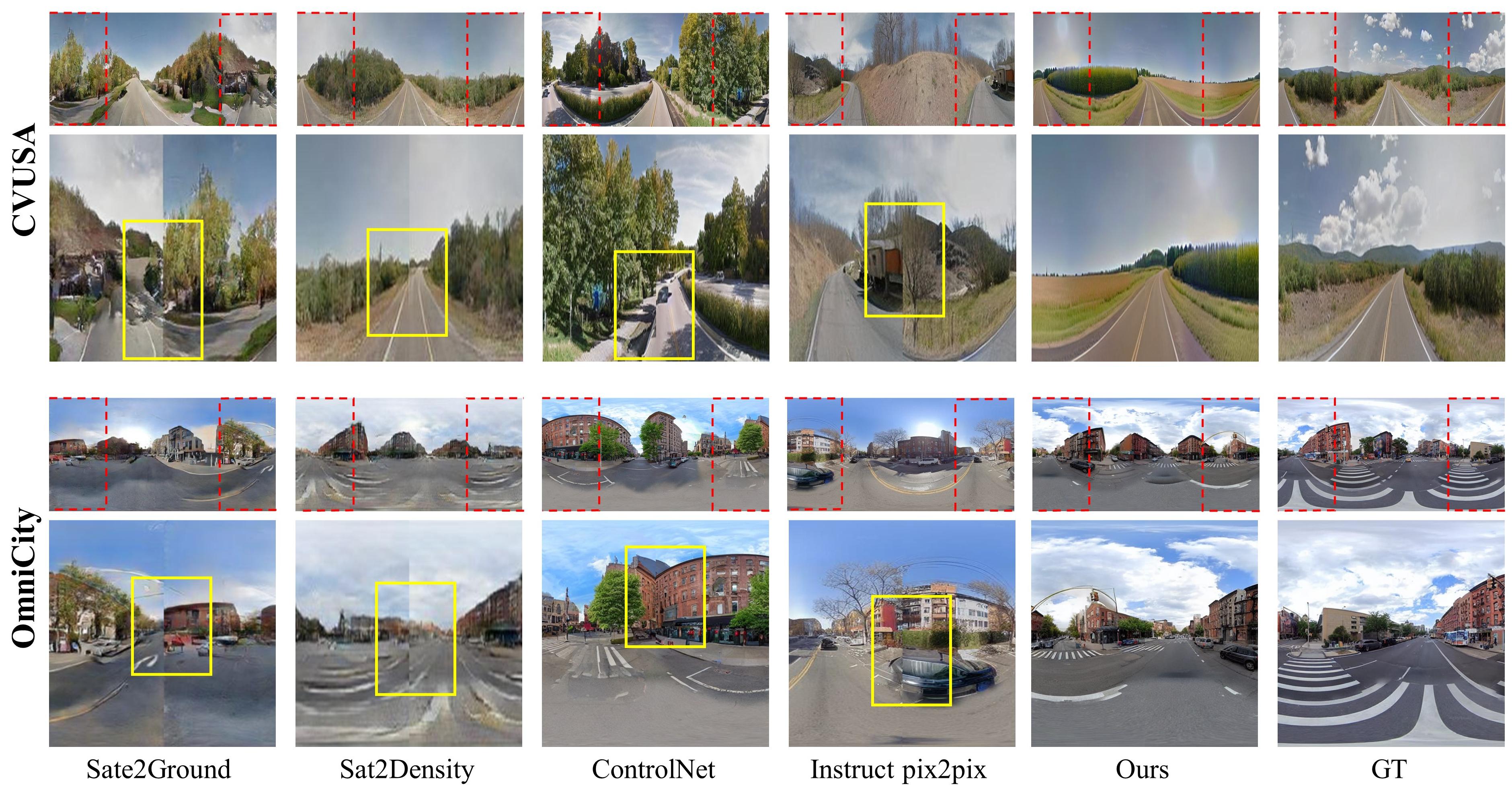}
    \caption{Qualitative results of the panorama continuity evaluation on CVUSA \cite{Zhai_2017_CVPR} and Omnicity \cite{Li_2023_CVPR}, respectively. By stitching the right 90° of the synthesis panorama to the left side of the image, our method demonstrates excellent consistency in texture and structure compared to other methods.}
    \label{Consistency}
\end{figure*}

For street-view panorama synthesis, another important evaluation factor is the continuity between the left and right sides of the image.
As illustrated in the qualitative results in Figure \ref{Consistency}, both GAN-based and diffusion-based methods produce synthesis results with apparent boundary lines, as they treat panorama synthesis as a general image synthesis task.
In contrast, our method constructs structural controls from a continuous scene composed of 3D voxels projected onto panoramic street views, allowing seamless integration at the left and right boundaries. 
For texture controls, the texture mapping features at the left and right positions of the street views are derived from proximate and continuous positions on the satellite image. Owing to these continuous structural and textural constraints, our method produces panoramic images with excellent 360° coherence.

\subsection{Ablation Study}

\begin{figure*}[!h]
    \includegraphics[width=\linewidth]{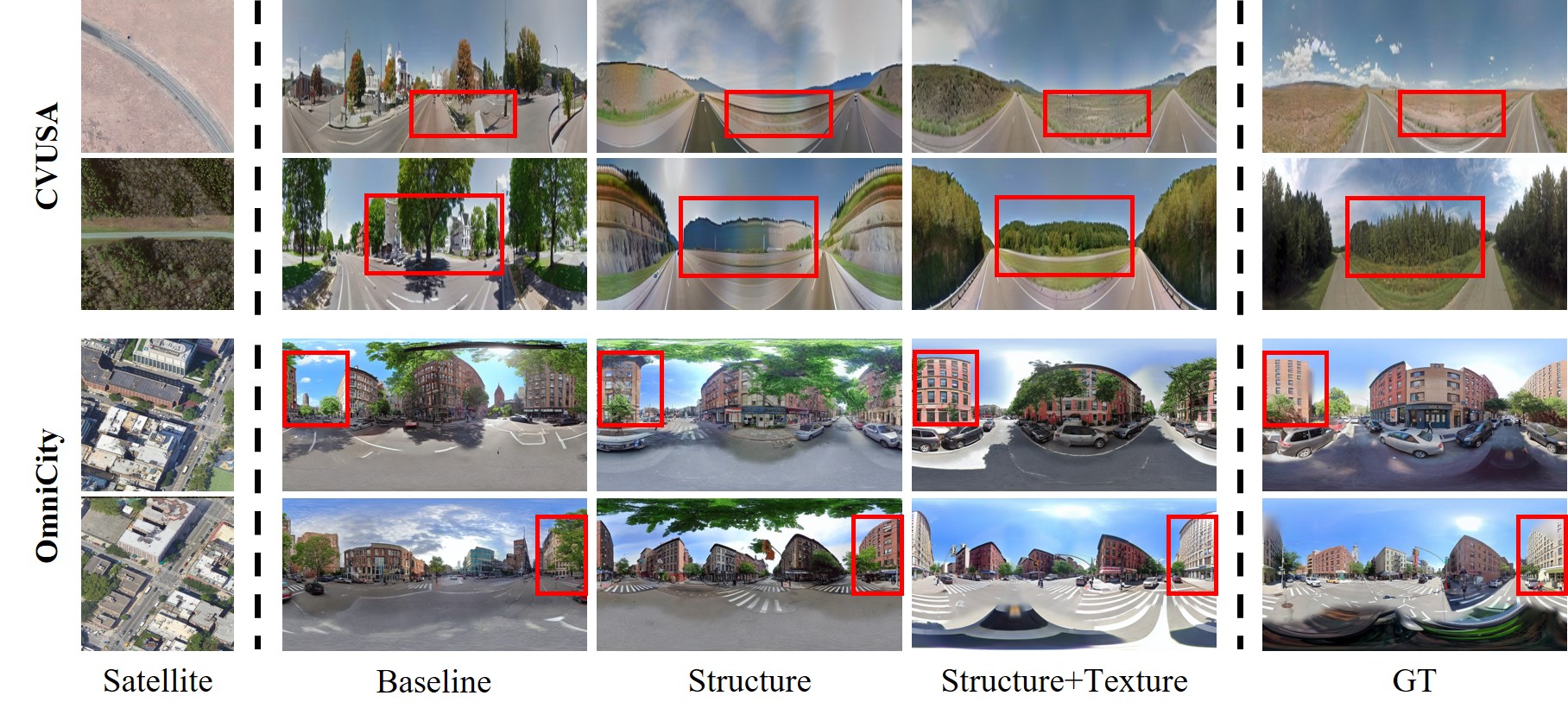}
    \caption{Qualitative ablation results on CVUSA \cite{Zhai_2017_CVPR} and OmniCity \cite{Li_2023_CVPR}. In the synthesis results, the first column represents the baseline without any structure or texture controls, the second column represents using only structure constraints, and the third column represents using both structure and texture (w/ CVTM) controls.}
    \label{fig:ablation0}
\end{figure*}

In our ablation study, we first assessed the effectiveness of our structure and texture control modules. As shown in the first two rows of each dataset of Table \ref{tab:ablation0}, using structural information derived from satellites as input proved effective, achieving improvements across multiple metrics such as SSIM \cite{SSIM2004}, FID \cite{FID}, and KID \cite{KID}. 
The last two rows of each dataset show the results of using direct Cross-Attention to incorporate global textures (w/o CVTM) and our Cross-View Texture Mapping (w/ CVTM) methods. Compared to the direct incorporation global textures, the approach guided by cross-view mapping relationships effectively assigns local textures from corresponding satellite regions to the appropriate locations in street-view images.
Figure \ref{fig:ablation0} presents qualitative ablation results on CVUSA \cite{Zhai_2017_CVPR} and OmniCity \cite{Li_2023_CVPR}, where structural control contributes to consistent content distribution, and texture control enhances the consistency of generated textures in buildings and forests.

\begin{table*}[!ht]
\centering
\caption{Quantitative ablation for different types of controls on CVUSA \cite{Zhai_2017_CVPR} and OmniCity \cite{Li_2023_CVPR}, including Structure, Texture (w/o CVTM), and Texture (w/ CVTM).}
\resizebox{\textwidth}{!}{%
\begin{tabular}{l|ccc|ccccc}
\toprule
Datasets & Structure & Texture (w/o CVTM) & Texture (w/ CVTM) & SSIM ($\uparrow$) & SD ($\uparrow$) & PSNR ($\uparrow$) & FID ($\downarrow$) & KID ($\downarrow$) \\
\midrule
\multirow{4}{*}{CVUSA} &  &  & & 0.277 & 15.22 & 11.182 & 44.63 & 0.044 \\
 & \checkmark &  &  & 0.312 & 15.30 & 10.358 & 41.19 & 0.039 \\
 & \checkmark & \checkmark &  & 0.283 & 15.65 & 10.913 & 33.51 & 0.020 \\
 & \checkmark &  & \checkmark & \textbf{0.371} & \textbf{16.31} & \textbf{12.000} & \textbf{23.67} & \textbf{0.018} \\
\midrule
\multirow{4}{*}{OmniCity} &  &  &  & 0.297 & 14.64 & 10.703 & 59.99 & 0.056 \\
 & \checkmark &  &  & 0.309 & 14.54 & \textbf{11.417} & 43.06 & 0.042 \\
 & \checkmark & \checkmark &  & 0.345 & 14.73 & 10.899 & 64.33 & 0.059 \\
 & \checkmark &  & \checkmark & \textbf{0.353} & \textbf{15.17} & 11.127 & \textbf{42.01} & \textbf{0.033} \\
\bottomrule
\end{tabular}
}
\label{tab:ablation0}
\end{table*}

\begin{table*}[!ht]
\centering
\caption{
Ablation results for varying depth estimations on CVUSA \cite{Zhai_2017_CVPR} and OmniCity \cite{Li_2023_CVPR} datasets. The impact of adjusted depth results on experimental metrics is minimal.
}
\scalebox{0.76}{%
    \begin{tabular}{l|ccccc|ccccc}
    \toprule
    \multirow{2}{*}{Method} & \multicolumn{5}{c|}{OmniCity}  & \multicolumn{5}{c}{CVUSA}\\
    \cmidrule(lr){2-6} \cmidrule(lr){7-11}
     & SSIM ($\uparrow$)  & SD ($\uparrow$)  & PSNR ($\uparrow$) & FID ($\downarrow$) & KID ($\downarrow$)  & SSIM ($\uparrow$) & SD ($\uparrow$) & PSNR ($\uparrow$)  & FID ($\downarrow$) & KID ($\downarrow$) \\
    \midrule
    Ours ($\times$ 0.9) & 0.350 & 15.10 & 11.111 & 43.58  & 0.034 & 0.365 & 16.30  & 11.943 & 24.11 & 0.019 \\
    Ours ($\times$ 1.1)  & 0.349 & 15.11 & 11.104  & 44.76 
     & 0.037  & 0.368 & 16.29 & 11.950 & \textbf{23.13} & 0.019 \\
    Ours  & \textbf{0.353} & \textbf{15.17}  & \textbf{11.127}   & \textbf{42.01} & \textbf{0.033} & \textbf{0.371} & \textbf{16.31} & \textbf{12.000} & 23.67 & \textbf{0.018} \\
    $\Delta$ & 0.004 & 0.07 & 0.023 & 2.75 & 0.004 & 0.006 & 0.02 & 0.057 & 0.98 & 0.001 \\
    \bottomrule
    \end{tabular}}
\label{tab:ablation1}
\end{table*}

Additionally, as the intermediaries for constructing both structural and textural controls, the 3D voxels derived from satellite depth estimation results significantly impact the accuracy of cross-view controls. Therefore, the precision of satellite depth estimation directly influences the effectiveness of these controls.
To simulate depth estimation inaccuracies, we apply scaling factors (0.9 and 1.1) to the depth estimation results before generating street-view images, as detailed in Table \ref{tab:ablation1}. The experimental results indicate that while our method relies on depth estimation, the stability of the model's output remains high, with minimal fluctuation in performance metrics.

\subsection{Experimental results using additional data sources}

In this section, we provide more experimental results of real-wolrd application scenarios using additional data sources.
In addition to the satellite images, other inputs such as textual data, building height data, and public map data (e.g. OpenStreetMap\footnote{\url{https://www.openstreetmap.org/}}) can also be used for generating street-view images. 
In this study, we explored the synthesis of street-view images using multiple data sources on the OmniCity \cite{Li_2023_CVPR} dataset and analyzed their impacts. Based on OmniCity street-view images, we generate corresponding text prompts of street-views images using the CLIP \cite{radford2021learning} model, and supplement the corresponding historical satellite imagery and OSM map data based on the street view capture locations.

\begin{figure*}[!h]
    \includegraphics[width=\linewidth]{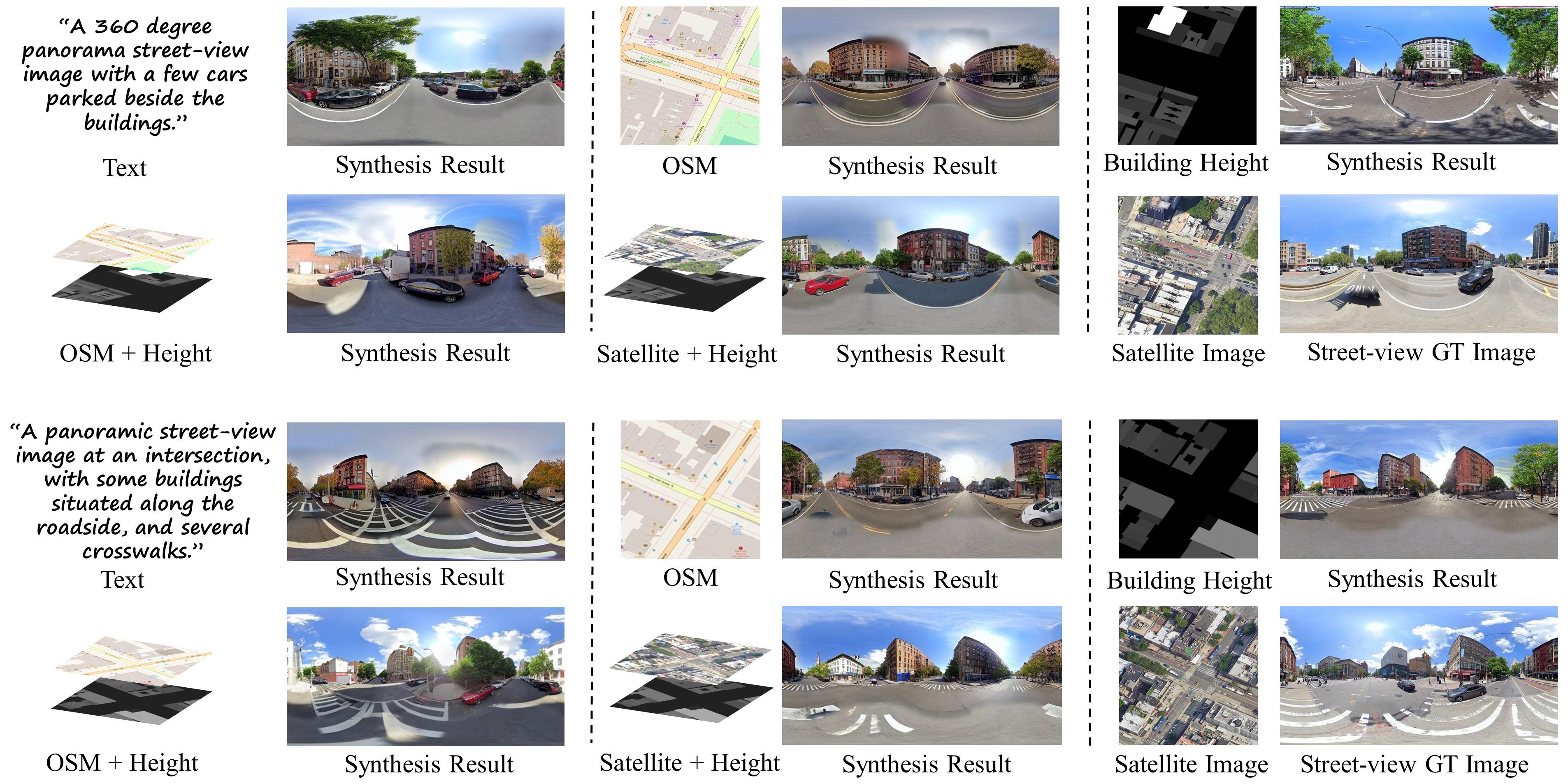}
    \caption{Qualitative comparison of different input types on the OmniCity \cite{Li_2023_CVPR} dataset. Using satellite image and building height as input achieves the best
results in all cases.
    }
    \label{fig:mutilsource}
\end{figure*}

As shown in Figure \ref{fig:mutilsource}, textual data can provide some global information about the scene , but its lack of detail and specificity results in visually unrealistic images. OSM (OpenStreetMap) data offers semantic features of different areas, such as roads, buildings, and parks. These semantic features aid in generating street-view images with consistent semantic content. However, when using only OSM data, the structure and texture of the synthesized street view images still show a certain gap compared to real images.
Building height data provides the outlines of buildings, and street-view images synthesized using this data show consistent building contours but lack texture and detail. 
Combining OSM and building height data for street view synthesis perform well in terms of semantics and structure. However, there are still deficiencies in texture details, such as building colors.
Combining satellite imagery and building height data yields street-view images that are optimal in both structure and texture, visually closest to real street views.
Table \ref{tab:mutil} provide the quantitative results obtained from different types of input data. Due to the rich texture information in satellite images, our CrossViewDiff achieved SSIM \cite{SSIM2004} and FID \cite{FID} scores of 0.361 and 37.89, respectively, representing improvements of 4.6\% and 17.6\% compared to the results synthesized using OSM and building height data as inputs.

\begin{table*}[ht]
\centering
\caption{
Quantitative comparison of different types of input data on the OmniCity dataset. Using satellite image and building height as input data achieves optimal performance, with a significant improvements compared with other input cases.
}

\scalebox{0.9}{%
\begin{tabular}{l|ccccc}

\toprule
 Input data & SSIM ($\uparrow$)  & SD ($\uparrow$)  & PSNR ($\uparrow$)& FID ($\downarrow$) & KID ($\downarrow$) \\
\midrule
Text & 0.298 & 14.54 & 11.131 & 82.37   & 0.069 \\
OSM  & 0.294 & 14.67 & 10.741  & 43.26 & 0.034 \\
Building height & 0.327 & 14.65  & 11.422   & 47.94 & 0.044  \\
OSM + Building height  & 0.345  & 14.79 & \textbf{11.748} & 45.98  & 0.039 \\
Satellite + Building height & \textbf{0.361} & \textbf{15.21}  & 11.512 & \textbf{37.89} & \textbf{0.027} \\
\bottomrule
\end{tabular}
}
\label{tab:mutil}
\end{table*}

\vspace{-10pt}

\begin{figure*}[h]
    \includegraphics[width=\linewidth]{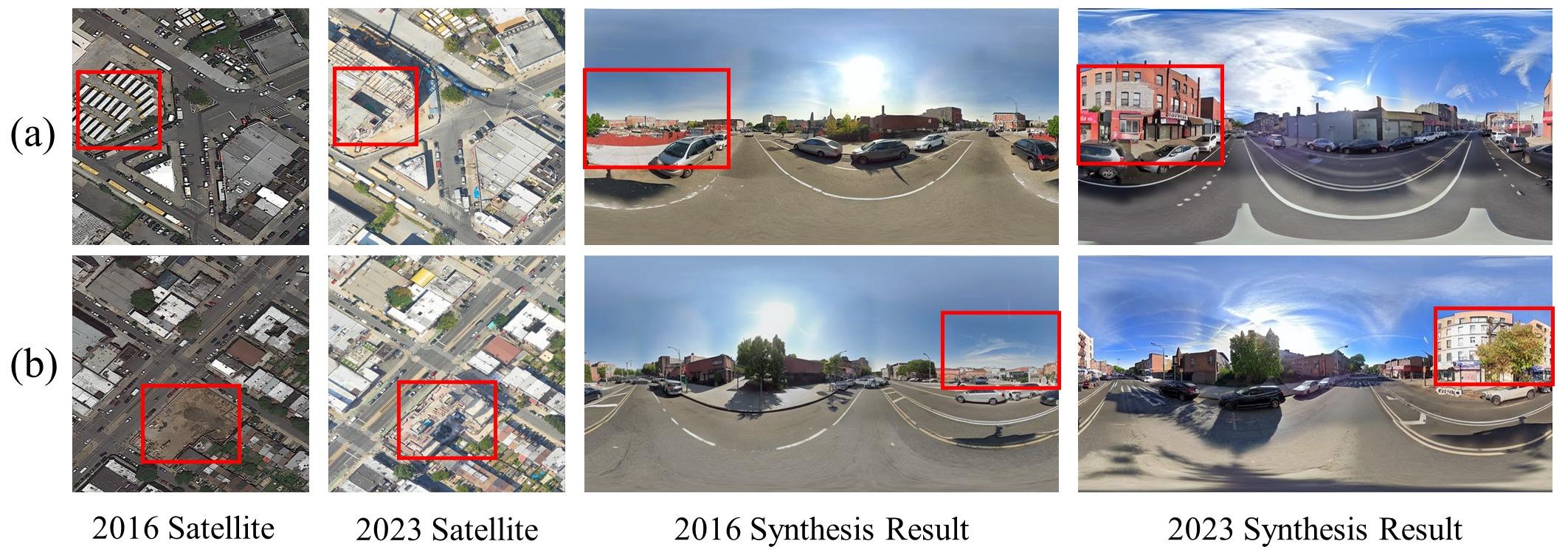}
    \caption{Visualization results of street-view synthesis from satellite images taken at different times. The areas highlighted in red indicate regions where terrain changes have occurred over time.}
    \label{fig:History}
\end{figure*}


Next, we explored the results of synthesizing street-view images using satellite imagery data from different years. As shown in Figure \ref{fig:History}, significant changes in terrain features over time can also be observed in our synthesized street-view images, such as the transformation of parking lots or vacant lots into buildings within the areas highlighted in red. 
Given the relatively recent widespread adoption of street-view imaging compared to the earlier availability of remote sensing satellite imagery, our effective satellite-to-street-view synthesis method unveils historical scenes from earlier times, offering practical application value.

\subsection{Limitation analysis}

\begin{figure*}[ht]
\centering
    \includegraphics[width=1\linewidth]{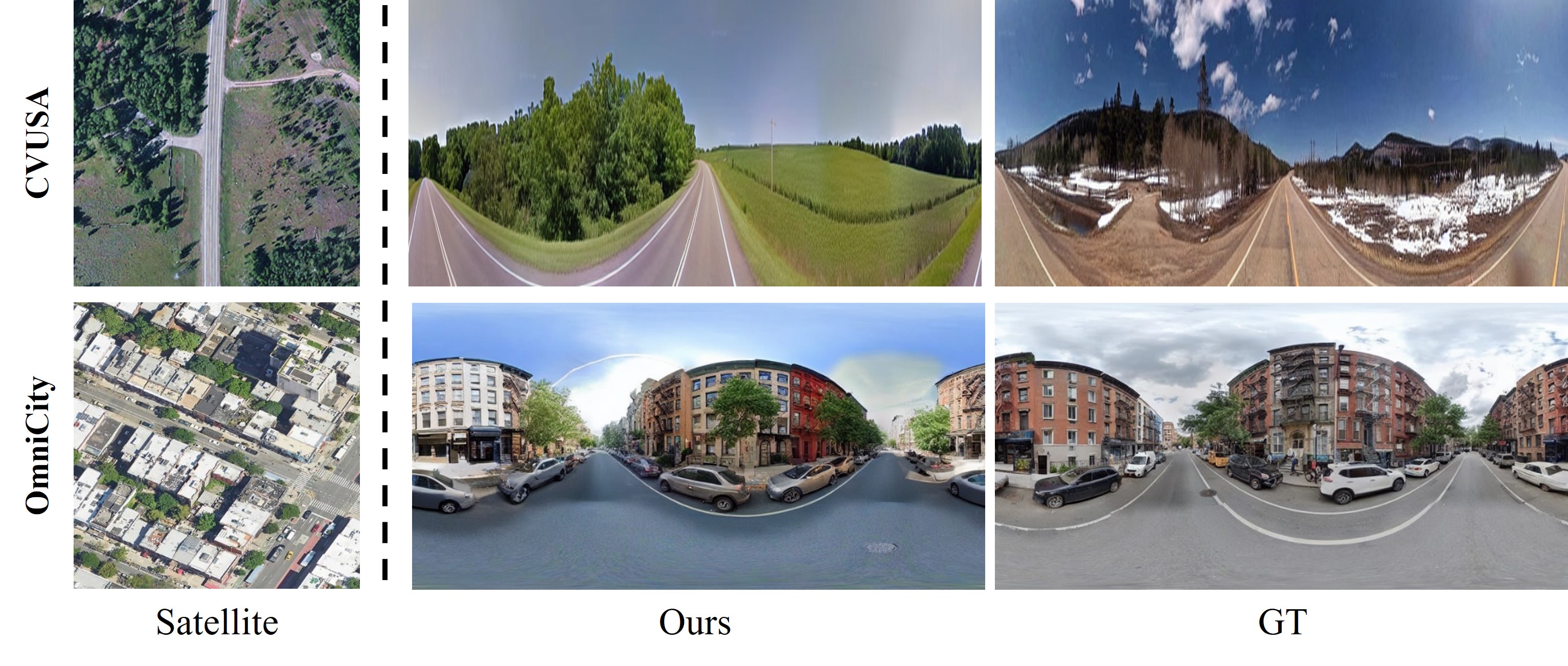}
    \caption{Typical failure cases of our method. 
    The first row of images shows that as the satellite and street-view images provided in the dataset were not taken at the same season, the synthetic image may not be consistent with the ground truth even if it is consistent with the satellite's features.
    The second row shows a significant discrepancy in the sky areas of the synthesized street views, as sky region information cannot be obtained from satellite images. 
    Additionally, vehicles and other moving objects pose significant challenges to cross-view synthesis.
    }
    \label{fig:Limitations}
\end{figure*}

Despite the above advantages, street-view images generated by CrossViewDiff still have several limitations. Although we fused features rich in structural and textural information based on satellite image, the gap between the two viewpoints is still large, and Stable Diffusion is more capable of creating additional details that do not actually exist. 
Figure \ref{fig:Limitations} provides some typical failure cases obtained by CrossViewDiff.
For satellite and street-view images that were not taken at the same season, even though the synthetic street-view image is consistent with the satellite's features, it may not be consistent with the ground truth.
Besides, in less constrained regions of the image such as the sky, the synthesis result is somewhat different from GT and has a certain amount of color shifting, resulting in the relatively low PSNR to some extent.
Moreover, due to the presence of moving objects such as pedestrians and vehicles in the scene, achieving consistency in cross-view synthesis results remains challenging.

\section{Conclusion}
\label{sec:conclusion}
In this work, we have proposed CrossViewDiff, a cross-view diffusion model to synthesize a street-view panorama from a given satellite image. 
The core of our diffusion model is a cross-view control guided denoising process that incorporates the structure and texture controls constructed by satellite scene structure estimation and cross-view texture mapping via an enhanced cross-view attention module.
Qualitative and quantitative results show that our method generates street-view panoramas with better consistency and perceptual quality as well as more realistic structures and textures compared with the state-of-the-art.
We believe that this paper motivates new ideas and inspirations for large-scale city simulation and 3D scene reconstruction.
In our future work, we will further explore the fusion of more types of multimodal data including textual data, map data, 3D data, and multi-temporal satellite imagery to enhance the quality and realism of the synthesized street-view images. 
We also plan to extend our method to more cities and improve our methods for more complex application scenes such as urban planning, virtual tourism, and intelligent navigation.

\section*{Declarations}
\label{sec:declaration}

\textbf{Data Availability} The datasets used this study can be accessed from: 1) CVUSA: \url{https://mvrl.cse.wustl.edu/datasets/cvusa}. 2) CVACT: \url{https://github.com/Liumouliu/OriCNN}. 3) OmniCity: \url{https://city-super.github.io/omnicity}.

\textbf{Code Availability} The implementation code and models related to the paper will be released at \url{https://opendatalab.github.io/CrossViewDiff}.

\FloatBarrier

\bibliography{reference}
\bibliographystyle{iclr2025_conference}

\end{document}